\documentclass[10pt,twocolumn,letterpaper]{article}

\usepackage{cvpr}
\usepackage{times}
\usepackage{epsfig}
\usepackage{graphicx}
\usepackage{amsmath}
\usepackage{amssymb}
\usepackage{breqn}
\usepackage{algorithm, algorithmic}
\usepackage{setspace}

\usepackage[pagebackref=true,breaklinks=true,letterpaper=true,colorlinks,bookmarks=false]{hyperref}

\cvprfinalcopy 

\def\cvprPaperID{919} 

\ifcvprfinal\fi
\begin{document}

\title{Detecting Oriented Text in Natural Images by Linking Segments}

\author{Baoguang Shi$^{1}$ \quad Xiang Bai$^{1}$\thanks{Corresponding author.} \quad Serge Belongie$^{2}$\\
$^{1}$School of EIC, Huazhong University of Science and Technology\\
$^{2}$Department of Computer Science, Cornell Tech\\
{\tt\small shibaoguang@gmail.com \quad xbai@hust.edu.cn \quad sjb344@cornell.edu}
}

\maketitle
\thispagestyle{empty}

\begin{abstract}
Most state-of-the-art text detection methods are specific to horizontal Latin text and are not fast enough for real-time applications.
We introduce Segment Linking (SegLink), an oriented text detection method.
The main idea is to decompose text into two locally detectable elements, namely segments and links.
A segment is an oriented box covering a part of a word or text line;
A link connects two adjacent segments, indicating that they belong to the same word or text line.
Both elements are detected densely at multiple scales by an end-to-end trained, fully-convolutional neural network.
Final detections are produced by combining segments connected by links.
Compared with previous methods, SegLink improves along the dimensions of accuracy, speed, and ease of training.
It achieves an f-measure of 75.0\% on the standard ICDAR 2015 \emph{Incidental} (Challenge 4) benchmark, outperforming the previous best by a large margin.
It runs at over 20 FPS on 512$\times$512 images.
Moreover, without modification, SegLink is able to detect long lines of non-Latin text, such as Chinese.
\end{abstract}

\section{Introduction}

\begin{figure}[t]
  \centering
  \includegraphics[width=\linewidth]{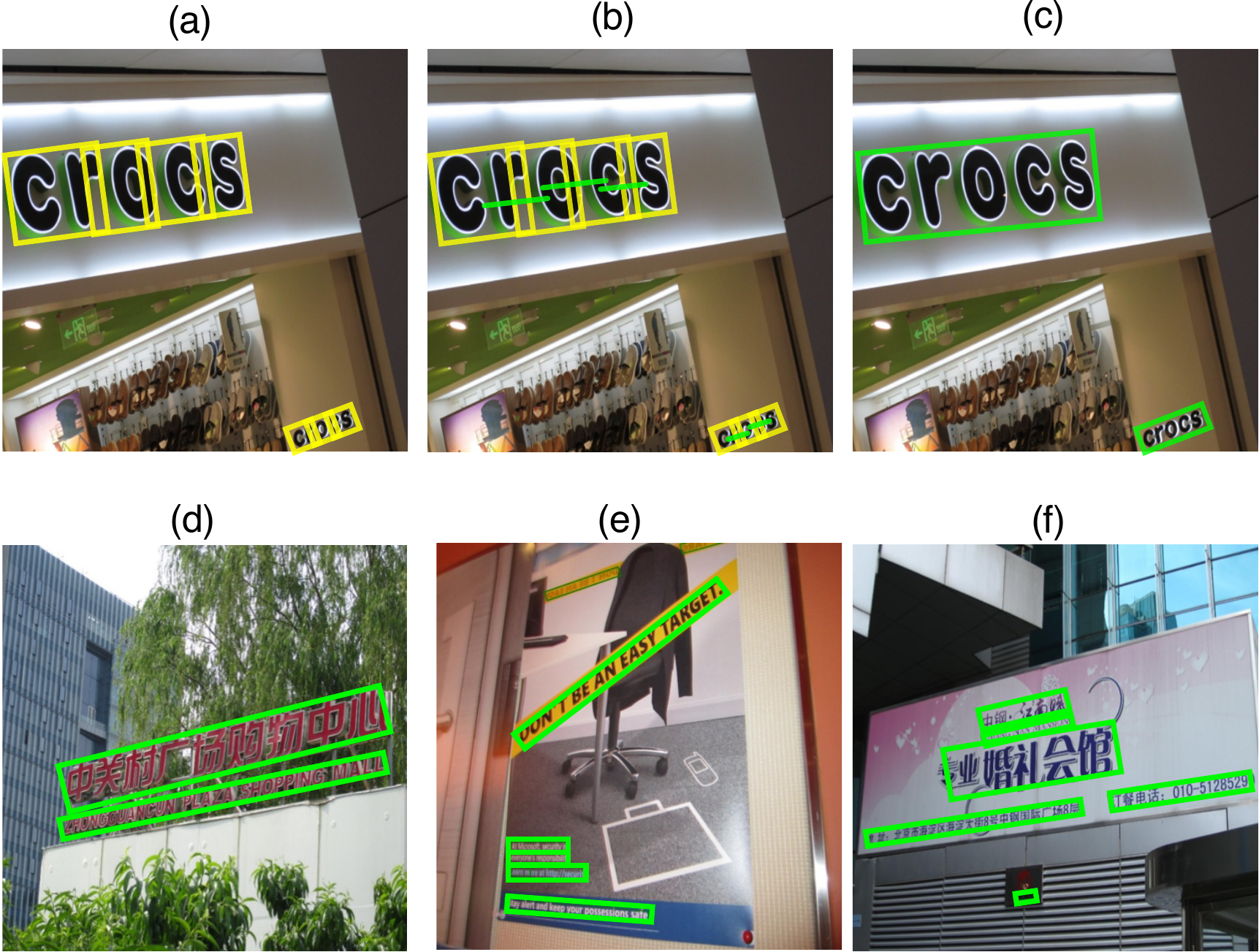}
  \caption{\textbf{SegLink Overview.} The upper row shows an image with two words of different scales and orientations. (a) Segments (yellow boxes) are detected on the image. (b) Links (green lines) are detected between pairs of adjacent segments. (c) Segments connected by links are combined into whole words. (d-f) SegLink is able to detect long lines of Latin and non-Latin text, such as Chinese.}
  \label{fig:seg-link}
\end{figure}

\begin{figure*}[t]
  \centering
  \includegraphics[width=0.85\linewidth]{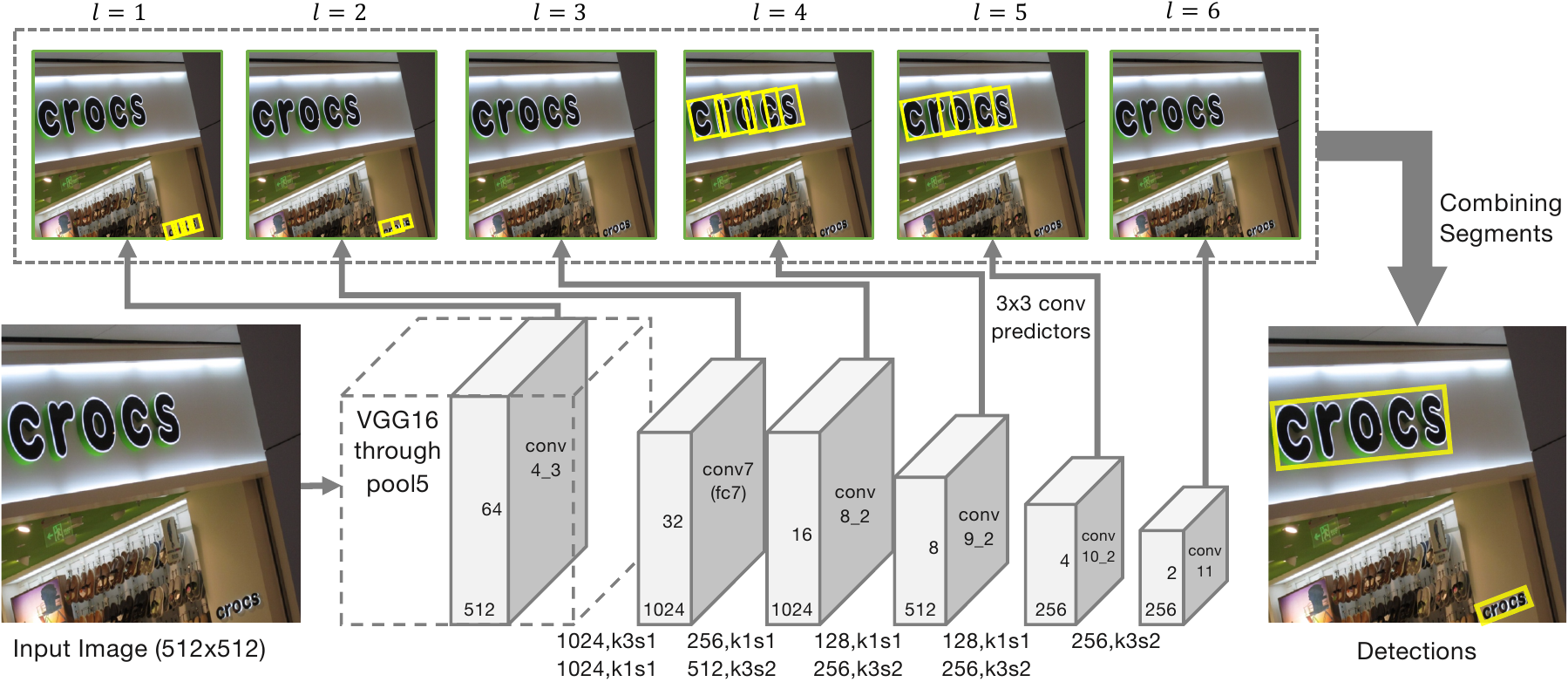}
  \caption{\textbf{Network Architecture.} The network consists of convolutional feature layers (shown as gray blocks) and convolutional predictors (thin gray arrows). Convolutional filters are specified in the format of ``(\#filters),k(kernel size)s(stride)''. A multi-line filter specification means a hidden layer between. Segments (yellow boxes) and links (not displayed) are detected by convolutional predictors on multiple feature layers (indexed by $l=1\dots6$) and combined into whole words by a combining algorithm.}
  \label{fig:model-overview}
\end{figure*}

Reading text in natural images is a challenging task under active research. It is driven by many real-world applications, such as Photo OCR~\cite{iccv/BissaccoCNN13}, geo-location, and image retrieval~\cite{ijcv/JaderbergSVZ16}.
In a text reading system, text detection, \emph{i.e.} localizing text with bounding boxes of words or text lines, is usually the first step of great significance.
In a sense, text detection can be seen as object detection applied to text, where words/characters/text lines are taken as the detection targets.
Owing to this, a new trend has emerged recently that state-of-the-art text detection methods~\cite{ijcv/JaderbergSVZ16, cvpr/GuptaVZ16, eccv/TianHHH016, cvpr/ZhangZSYLB16} are heavily based on the advanced general object detection or segmentation techniques, \emph{e.g.}~\cite{iccv/Girshick15, cvpr/GirshickDDM14, cvpr/LongSD15}.

Despite the great success of the previous work, we argue that the general detection methods are not well suited for text detection, for two main reasons.
First, word/text line bounding boxes have much larger aspect ratios than those of general objects. An (fast/faster) R-CNN~\cite{cvpr/GirshickDDM14,iccv/Girshick15,nips/RenHGS15}- or SSD~\cite{eccv/LiuAESRFB16}-style detector may suffer from the difficulty of producing such boxes, owing to its proposal or anchor box design.
In addition, some non-Latin text does not have blank spaces between words, hence the even larger bounding box aspect ratios, which make the problem worse.
Second, unlike general objects, text usually has a clear definition of orientation~\cite{cvpr/YaoBLMT12}.
It is important for a text detector to produce oriented boxes.
However, most general object detection methods are not designed to produce oriented boxes.

To overcome the above challenges, we tackle the text detection problem in a new perspective. We propose to decompose long text into two smaller and locally-detectable elements, namely \emph{segment} and \emph{link}.
As illustrated in Fig.~\ref{fig:seg-link}, a segment is an oriented box that covers a part of a word (for clarity we use ``word'' here and later on, but segments also work seamlessly on text lines that comprise multiple words); A link connects a pair of adjacent segments, indicating that they belong to the same word. Under the above definitions, a word is located by a number of segments with links between them.
During detection, segments and links are densely detected on an input image by a convolutional neural network. Then, the segments are combined into whole words according to the links.

The key advantage of this approach is that long and oriented text is now detected locally since both basic elements are \emph{locally-detectable}: Detecting a segment does not require the whole word to be observed. And neither does a link since the connection of two segments can be inferred from a local context.
Thereafter, we can detect text of any length and orientation with great flexibility and efficiency.

Concretely, we propose a convolutional neural network (CNN) model to detect both segments and links simultaneously, in a \emph{fully-convolutional} manner.
The network uses VGG-16~\cite{corr/SimonyanZ14a} as its backbone. A few extra feature layers are added onto it.
Convolutional predictors are added to 6 of the feature layers to detect segments and links at different scales.
To deal with redundant detections, we introduce two types of links, namely \emph{within-layer links} and \emph{cross-layer links}.
A within-layer link connects a segment to its neighbors on the same layer.
A cross-layer link, on the other hand, connects a segment to its neighbors on the lower layer.
In this way, we connect segments of adjacent locations as well as scales.
Finally, we find connected segments with a depth-first search (DFS) algorithm and combine them into whole words.

Our main contribution is the novel segment-linking detection method. Through experiments, we show that the proposed method possesses several distinctive advantages over the other state-of-the-art methods: 1)~\emph{Robustness}: SegLink models the structure of oriented text in a simple and elegant way, with robustness against complex backgrounds. Our method achieves highly competitive results on standard datasets. In particular, it outperforms the previous best by a large margin in terms of f-measure (75.0\% vs 64.8\%) on the ICDAR 2015 Incidental (Challenge 4) benchmark~\cite{icdar/KaratzasGNGBIMN15}; 2)~\textit{Efficiency}: SegLink is highly efficient due to its single-pass, fully-convolutional design. It processes more than 20 images of 512x512 size per second; 3)~\textit{Generality}: Without modification, SegLink is able to detect long lines of non-Latin text, such as Chinese. We demonstrate this capability on a multi-lingual dataset.

\section{Related Work}

\paragraph{Text Detection}
Over the past few years, much research effort has been devoted to the text detection problem~\cite{icpr/WangWCN12,eccv/WangB10,pami/NeumannM16,pami/NeumannM16,cvpr/YaoBLMT12,iccv/HuangLYW13,eccv/HuangQT14,cvpr/ZhangZSYLB16,cvpr/ZhangSYB15,iccv/BissaccoCNN13,ijcv/JaderbergSVZ16,cvpr/GuptaVZ16,eccv/TianHHH016,corr/YaoBSZZC16}. Based on the basic detection targets, the previous methods can be roughly divided into three categories: \emph{character-based}, \emph{word-based} and \emph{line-based}.
Character-based methods~\cite{pami/NeumannM16,eccv/WangB10, icpr/WangWCN12, eccv/JaderbergVZ14, iccv/HuangLYW13, eccv/HuangQT14} detect individual characters and group them into words.
These methods find characters by classifying candidate regions extracted by region extraction algorithms or by classifying sliding windows.
Such methods often involve a post-processing step of grouping characters into words.
Word-based methods~\cite{ijcv/JaderbergSVZ16, cvpr/GuptaVZ16} directly detect word bounding boxes.
They often have a similar pipeline to the recent CNN-based general object detection networks.
Though achieving excellent detection accuracies, these methods may suffer from performance drop when applied to some non-Latin text such as Chinese, as we mentioned earlier.
Line-based methods~\cite{cvpr/ZhangSYB15, cvpr/ZhangZSYLB16,corr/YaoBSZZC16} find text regions using some image segmentation algorithms. They also require a sophisticated post-processing step of word partitioning and/or false positive removal.
Compared with the previous approaches, our method predicts segments and links jointly in a single forward network pass.
The pipeline is much simpler and cleaner.
Moreover, the network is end-to-end trainable.

Our method is similar in spirit to a recent work~\cite{eccv/TianHHH016}, which detects text lines by finding and grouping a sequence of \emph{fine-scale text proposals} through a CNN coupled with recurrent neural layers.
In contrast, we detect oriented segments only using convolutional layers, yielding better flexibility and faster speed.
Also, we detect links explicitly using the same strong CNN features for segments, improving the robustness.

\paragraph{Object Detection}
Text detection can be seen as a particular instance of general object detection, which is a fundamental problem in computer vision.
Most state-of-the-art detection systems either classify some class-agnostic object proposals with CNN~\cite{cvpr/GirshickDDM14,iccv/Girshick15,nips/RenHGS15} or directly regress object bounding boxes from a set of preset boxes (\emph{e.g.} anchor boxes)~\cite{corr/RedmonDGF15,eccv/LiuAESRFB16}.

The architecture of our network inherits that of SSD~\cite{eccv/LiuAESRFB16}, a recent object detection model.
SSD proposed the idea of detecting objects on multiple feature layers with convolutional predictors.
Our model also detects segments and links in a very similar way.
Despite the model similarity, our detection strategy is drastically different: SSD directly outputs object bounding boxes. We, on the other hand, adopt a bottom-up approach by detecting the two comprising elements of a word or text line and combine them together.

\section{Segment Linking}

Our method detects text with a feed-forward CNN model. Given an input image $I$ of size $w_{I}\times h_{I}$, the model outputs a fixed number of segments and links, which are then filtered by their confidence scores and combined into whole word bounding boxes.
A bounding box is a rotated rectangle denoted by $b = (x_b, y_b, w_b, h_b, \theta_b)$, where $x_b, y_b$ are the coordinates of the center, $w_b, h_b$ the width and height, and $\theta_b$ the rotation angle.

\subsection{CNN Model}

Fig.~\ref{fig:model-overview} shows the network architecture.
Our network uses a pretrained VGG-16 network~\cite{corr/SimonyanZ14a} as its backbone (conv1 through pool5).
Following~\cite{eccv/LiuAESRFB16}, the fully-connected layers of VGG-16 are converted into convolutional layers (fc6 to conv6; fc7 to conv7).
They are followed by a few extra convolutional layers (conv8\_1 to conv11), which extract even deeper features with larger receptive fields.
Their configurations are specified in Fig.~\ref{fig:model-overview}.

Segments and links are detected on 6 of the feature layers, which are conv4\_3, conv7, conv8\_2, conv9\_2, conv10\_2, and conv11.
These feature layers provide high-quality deep features of different granularity (conv4\_3 the finest and conv11 the coarsest).
A convolutional predictor with $3 \times 3$ kernels is added to each of the 6 layers to detect segments and links.
We index the feature layers and the predictors by $l=1,\dots,6$.

\paragraph{Segment Detection}
Segments are also oriented boxes, denoted by $s=(x_{s},$ $y_{s},$ $w_{s},$ $h_{s},$ $\theta_{s})$.
We detect segments by estimating the confidence scores and geometric offsets to a set of \emph{default boxes}~\cite{eccv/LiuAESRFB16} on the input image.
Each default box is associated with a feature map location, and its score and offsets are predicted from the features at that location.
For simplicity, we only associate one default box with a feature map location.

Consider the $l$-th feature layer whose feature map size is $w_l \times h_l$.
A location $(x, y)$ on this map corresponds to a default box centered at $(x_{a}, y_{a})$ on the image, where
\begin{equation}
  x_{a} = \frac{w_I}{w_l}(x+0.5);\quad y_{a} = \frac{h_I}{h_l}(y+0.5)
\end{equation}
The width and height of the default box are both set to a constant $a_l$.

The convolutional predictor produces 7 channels for segment detection.
Among them, 2 channels are further softmax-normalized to get the segment score in $(0,1)$.
The rest 5 are the geometric offsets.
Considering a location $(x,y)$ on the map, we denote the vector at this location along the depth by $(\Delta x_{s},$ $\Delta y_{s},$ $\Delta w_{s},$ $\Delta h_{s},$ $\Delta \theta_{s})$.
Then, the segment at this location is calculated by:
\begin{align}
  x_{s} &= a_l \Delta x_{s} + x_{a} \label{eq:calc-segment-first} \\
  y_{s} &= a_l \Delta y_{s} + y_{a} \\
  w_{s} &= a_l \exp(\Delta w_{s}) \\
  h_{s} &= a_l \exp(\Delta h_{s}) \\
  \theta_{s} &= \Delta \theta_{s} \label{eq:calc-segment-last}
\end{align}
Here, the constant $a_l$ controls the scale of the output segments.
It should be chosen with regard to the receptive field size of the $l$-th layer.
We use an empirical equation for choosing this size: $a_l = \gamma \frac{w_{I}}{w_l}$, where $\gamma=1.5$.

\begin{figure}
  \centering
  \includegraphics[width=0.85\linewidth]{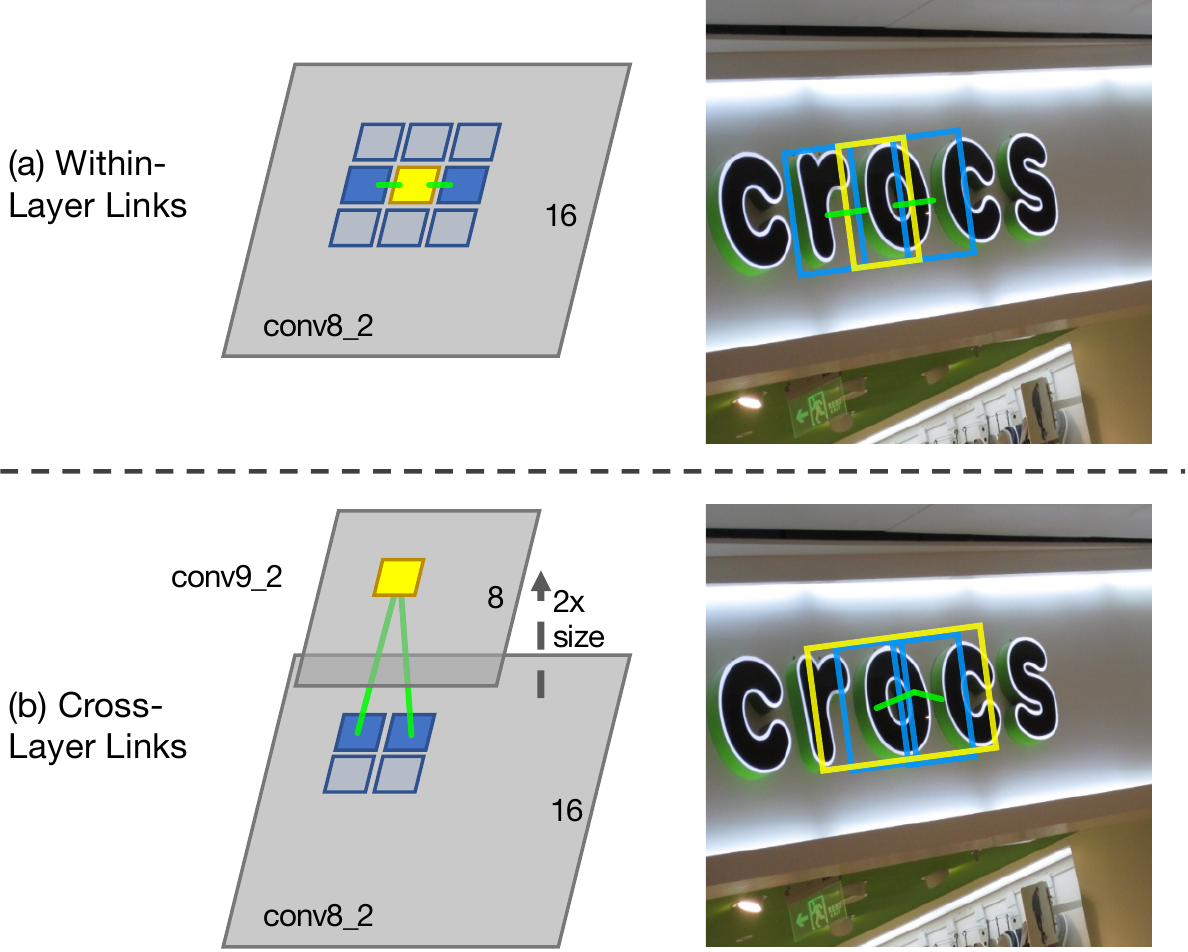}
  \caption{\textbf{Within-Layer and Cross-Layer Links.} (a) A location on conv8\_2 (yellow block) and its 8-connected neighbors (blue blocks with and without fill). The detected within-layer links (green lines) connect a segment (yellow box) and its two neighboring segments (blue boxes) on the same layer.
  (b) The cross-layer links connect a segment on conv9\_2 (yellow box) and two segments on conv8\_2 (blue boxes).}
  \label{fig:links}
\end{figure}

\paragraph{Within-Layer Link Detection}
A link connects a pair of adjacent segments, indicating that they belong to the same word.
Here, adjacent segments are those detected at adjacent feature map locations.
Links are not only necessary for combining segments into whole words but also helpful for separating two nearby words -- between two nearby words, the links should be predicted as negative.

We explicitly detect links between segments using the same features for detecting segments.
Since we detect only one segment at a feature map location, segments can be indexed by their map locations $(x,y)$ and layer indexes $l$, denoted by $s^{(x,y,l)}$. As illustrated in Fig.~\ref{fig:links}.a, we define the \emph{within-layer neighbors} of a segment as its 8-connected neighbors on the same feature layer:
\begin{dmath}
  {\cal N}^{w}_{s^{(x,y,l)}} = \{ s^{(x',y',l)} \}_{x-1\leq x' \leq x+1, y-1\leq y' \leq y+1} \setminus s^{(x,y,l)}
\end{dmath}
As segments are detected locally, a pair of neighboring segments are also adjacent on input image.
Links are also detected by the convolutional predictors.
A predictor outputs 16 channels for the links to the 8-connected neighboring segments.
Every 2 channels are softmax-normalized to get the score of a link.

\paragraph{Cross-Layer Link Detection}
In our network, segments are detected at different scales on different feature layers.
Each layer handles a range of scales.
We make these ranges overlap in order not to miss scales at their edges.
But as a result, segments of the same word could be detected on multiple layers at the same time, producing redundancies.

To address this problem, we further propose another type of links, called \emph{cross-layer links}.
A cross-layer link connects segments on two feature layers with adjacent indexes.
For example, cross-layer links are detected between conv4\_3 and conv7, because their indexes are $l=1$ and $l=2$ respectively.

An important property of such a pair is that the first layer always has twice the size as the second one, because of the down-sampling layer (max-pooling or stride-2 convolution) between them.
Note that this property only holds when all feature layers have even-numbered sizes.
In practice, we ensured this property by having the width and height of the input image both dividable by 128.
For example, an $1000\times 800$ image is resized to $1024 \times 768$, which is the nearest valid size.

As illustrated in Fig.~\ref{fig:links}.b, we define the \emph{cross-layer neighbors} of a segment as
\begin{dmath}
  {\cal N}^{c}_{s^{(x,y,l)}} = \{ s^{(x',y',l-1)} \}_{2x \leq x' \leq 2x+1, 2y\leq y' \leq 2y+1},
\end{dmath}
which are the segments on the preceeding layer.
Every segment has 4 cross-layer neighbors.
The correspondence is ensured by the double-size relationship between the two layers.

Again, cross-layer links are detected by the convolutional predictor.
The predictor outputs 8 channels for cross-layer links.
Every 2 channels are softmax-normalized to produce the score of a cross-layer link.
Cross-layer links are detected on feature layer $l=2\dots 6$, but not on $l=1$ (conv4\_3) since it has no preceeding feature layer.

With cross-layer links, segments of different scales can be connected and later combined.
Compared with the traditional non-maximum suppression,
cross-layer linking provides a trainable way of joining redundancies.
Besides, it fits seamlessly into our linking strategy and is easy to implement under our framework.

\begin{figure}[h]
  \centering
  \includegraphics[width=0.6\linewidth]{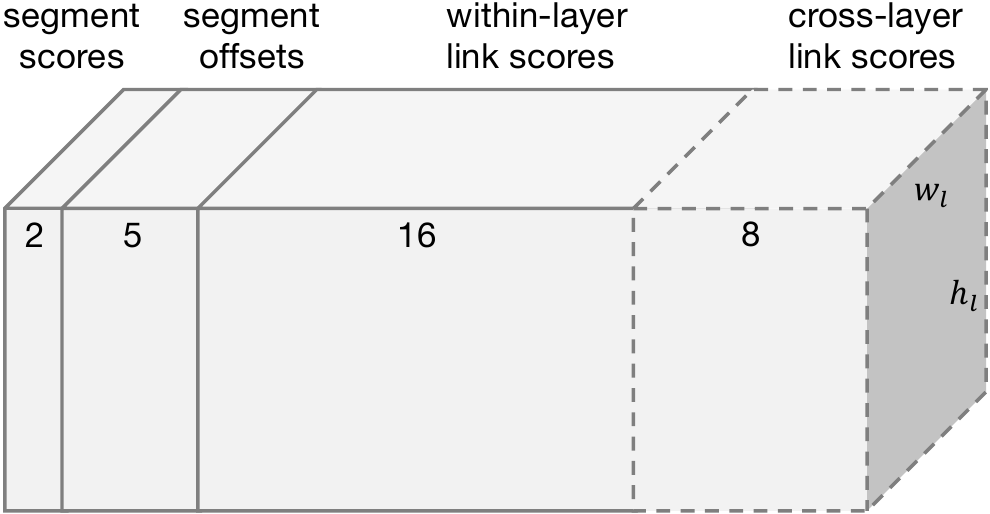}
  \caption{Output channels of a convolutional predictor. The block shows a $w_l \times h_l$ map of depth 31. The predictor of $l=1$ does not output the channels for corss-layer links.}
  \label{fig:predictor-output}
\end{figure}

\paragraph{Outputs of a Convolutional Predictor} Putting things together, Fig.~\ref{fig:predictor-output} shows the output channels of a convolutional predictor.
A predictor is implemented by a convolutional layer followed by some softmax layers that normalize the segment and link scores respectively.
Thereafter, all layers in our network are convolutional layers.
Our network is fully-convolutional.

\subsection{Combining Segments with Links}
After feed-forwarding, the network produces a number of segments and links (the number depends on the image size).
Before combination, the output segments and links are filtered by their confidence scores.
We set different filtering thresholds for segment and link, respectively $\alpha$ and $\beta$.
Empirically, the performance of our model is not very sensitive to these thresholds.
A 0.1 deviation on either thresholds from their optimal values results in less than 1\% f-measure drop.

Taking the filtered segments as nodes and the filtered links as edges, we construct a graph over them. Then, a depth-first search (DFS) is performed over the graph to find its connected components. Each component contains a set of segments that are connected by links.
Denoting a connected component by ${\cal B}$, segments within this component are combined following the procedures in Alg.~\ref{alg:combine-segments}.

\begin{algorithm}[h]
\caption{Combining Segments}
\setstretch{1.1}
\begin{algorithmic}[1]
  \STATE \textbf{Input:} ${\cal B} = \{ s^{(i)} \} _{i=1}^{|{\cal B}|}$ is a set of segments connected by links, where $s^{(i)} = (x_s^{(i)}, y_s^{(i)}, w_s^{(i)}, h_s^{(i)}, \theta_s^{(i)})$.
  \STATE Find the average angle $\theta_b := \frac{1}{|{\cal B}|} \sum_{{\cal B}} \theta_{s}^{(i)}$.
  \STATE For a straight line $(\tan \theta_b) x + b$, find the $b$ that minimizes the sum of distances to all segment centers $(x_s^{(i)}, y_s^{(i)})$.
  \STATE Find the perpendicular projections of all segment centers onto the straight line.
  \STATE From the projected points, find the two with the longest distance. Denote them by $(x_p, y_p)$ and $(x_q, y_q)$.
  \STATE $x_b := \frac{1}{2} (x_p + x_q)$
  \STATE $y_b := \frac{1}{2} (y_p + y_q)$
  \STATE $w_b := \sqrt{(x_p-x_q)^2+(y_p-y_q)^2} + \frac{1}{2} (w_p + w_q)$
  \STATE $h_b := \frac{1}{|{\cal B}|} \sum_{{\cal B}} h_{s}^{(i)}$
  \STATE $b := (x_b, y_b, w_b, h_b, \theta_b)$
  \STATE \textbf{Output:} $b$ is the combined bounding box.
\end{algorithmic}
\label{alg:combine-segments}
\end{algorithm}

\section{Training}

\subsection{Groundtruths of Segments and Links}

The network is trained by the direct supervision of groundtruth segments and links.
The groundtruths include the labels of all default boxes (\emph{i.e.} the label of their corresponding segments), their offsets to the default boxes, and the labels of all within- and cross-layer links.
We calculate them from the groundtruth word bounding boxes.

First, we assume that there is only one groundtruth word on the input image. A default box is labeled as positive iff 1) the center of the box is inside the word bounding box; 2) the ratio between the box size $a_l$ and the word height $h$ satisfies:
\begin{equation}
  \max (\frac{a_l}{h}, \frac{h}{a_l}) \leq 1.5
  \label{eq:anchor-word-ratio}
\end{equation}
Otherwise, the default box is labeled as negative.

Next, we consider the case of multiple words. A default box is labeled as negative if it does not meet the above-mentioned criteria for any word.
Otherwise, it is labeled as positive and matched to the word that has the closest size, \emph{i.e.} the one with the minimal value at the left-hand side of Eq.~\ref{eq:anchor-word-ratio}.

\begin{figure}[h]
  \centering
  \includegraphics[width=0.9\linewidth]{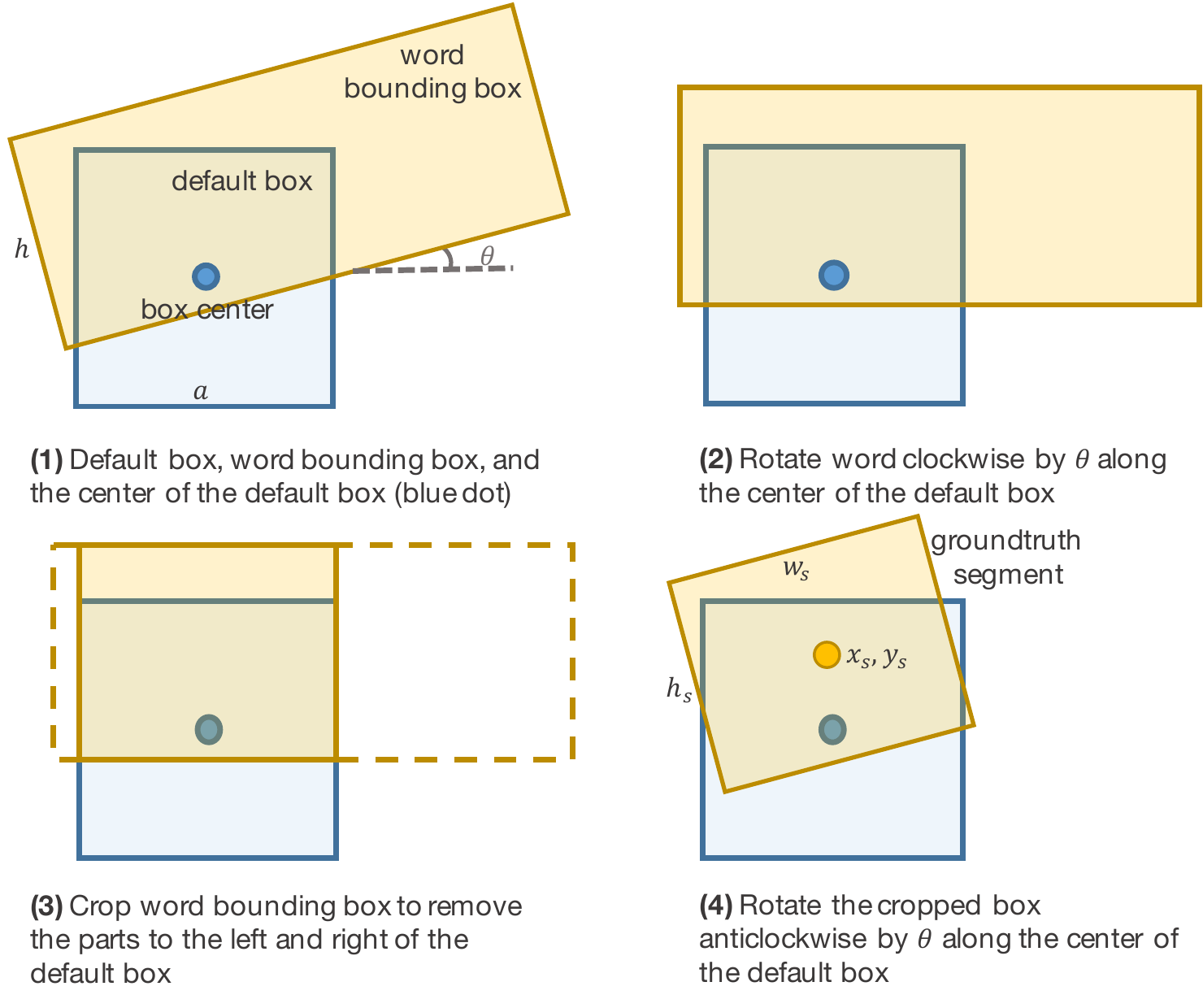}
  \caption{The steps of calculating a groundtruth segment given a default box and a word bounding box.}
  \label{fig:gt-segment}
\end{figure}

Offsets are calculated on positive default boxes.
First, we calculate the groundtruth segments following the steps illustrated in Fig.~\ref{fig:gt-segment}.
Then, we solve Eq.~\ref{eq:calc-segment-first} to Eq.~\ref{eq:calc-segment-last} to get the groundtruth offsets.

A link (either within-layer or cross-layer) is labeled as positive iff 1) both of the default boxes connected to it are labeled as positive; 2) the two default boxes are matched to the same word.

\subsection{Optimization}

\paragraph{Objective}
Our network model is trained by simultaneously minimizing the losses on segment classification, offsets regression, and link classification. Overall, the loss function is a weighted sum of the three losses:
\begin{dmath}
  L(\mathbf{y}_s, \mathbf{c}_s, \mathbf{y}_l, \mathbf{c}_l, \hat{\mathbf{s}}, \mathbf{s}) = \frac{1}{N_s} L_{\mathrm{conf}} (\mathbf{y}_s, \mathbf{c}_s)
      + \lambda_1 \frac{1}{N_s} L_{\mathrm{loc}} (\hat{\mathbf{s}}, \mathbf{s})
      + \lambda_2 \frac{1}{N_l} L_{\mathrm{conf}} (\mathbf{y}_l, \mathbf{c}_l)
\end{dmath}
Here, $\mathbf{y}_s$ is the labels of all segments. $\mathbf{y}_s^{(i)}=1$ if the $i$-th default box is labeled as positive, and $0$ otherwise. Likewise, $\mathbf{y}_l$ is the labels of the links.
$L_{\mathrm{conf}}$ is the softmax loss over the predicted segment and link scores, respectively $\mathbf{c}_s$ and $\mathbf{c}_l$. $L_{\mathrm{loc}}$ is the \emph{Smooth L1} regression loss~\cite{iccv/Girshick15} over the predicted segment geometries $\hat{\mathbf{s}}$ and the groundtruth $\mathbf{s}$.
The losses on segment classification and regression are normalized by $N_{s}$, which is the number of positive default boxes.
The loss on link classification is normalized by the number of positive links $N_{l}$.
The weight constants $\lambda_1$ and $\lambda_2$ are both set to $1$ in practice.

\paragraph{Online Hard Negative Mining}
For both segments and links, negatives take up most of the training samples.
Therefore, hard negative mining is necessary for balancing the positive and negative samples.
We follow the online hard negative mining strategy proposed in~\cite{cvpr/ShrivastavaGG16} to keep the ratio between the negatives and positives 3:1 at most.
Hard negative mining is performed separately for segments and links.

\paragraph{Data Augmentation}
We adopt an online augmentation pipeline that is similar to that of SSD~\cite{eccv/LiuAESRFB16} and YOLO~\cite{corr/RedmonDGF15}.
Training images are randomly cropped to a patch that has a minimum Jaccard overlap of $o$ with any groundtruth word
Crops are resized to the same size before loaded into a batch.
For oriented text, the augmentation is performed on the axis-aligned bounding boxes of the words.
The overlap $o$ is randomly chosen from $0$ (no constraint), $0.1$, $0.3$, $0.5$, $0.7$, and $0.9$ for every sample.
The crop size is randomly chosen from $[0.1, 1]$ of the original image size.
Training images are \emph{not} horizontally flipped.

\section{Experiments}

In this section, we evaluate the proposed method on three public datasets, namely ICDAR 2015 Incidental Text (Challenge 4), MSRA-TD500, and ICDAR 2013, using the standard evaluation protocol of each.

\subsection{Datasets}

\paragraph{SynthText in the Wild (SynthText)~\cite{cvpr/GuptaVZ16}} contains 800,000 \emph{synthetic} training images.
They are created by blending natural images with text rendered with random fonts, size, orientation, and color.
Text is rendered and aligned to carefully chosen image regions in order have a realistic look.
The dataset provides very detailed annotations for characters, words, and text lines.
We only use the dataset for pretraining our network.

\paragraph{ICDAR 2015 Incidental Text (IC15)~\cite{icdar/KaratzasGNGBIMN15}} is the Challenge 4 of the ICDAR 2015 Robust Reading Competition.
This challenge features \emph{incidental} scene text images taken by Google Glasses without taking care of positioning, image quality, and viewpoint.
Consequently, the dataset exhibits large variations in text orientation, scale, and resolution, making it much more difficult than previous ICDAR challenges.
The dataset contains 1000 training images and 500 testing images.
Annotations are provided as word quadrilaterals.

\paragraph{MSRA-TD500 (TD500)~\cite{cvpr/YaoBLMT12}} is the first standard dataset that focuses on oriented text.
The dataset is also multi-lingual, including both Chinese and English text.
The dataset consists of 300 training images and 200 testing images.
Different from IC15, TD500 is annotated at the level of text lines.

\paragraph{ICDAR 2013 (IC13)~\cite{icdar/KaratzasSUIBMMMAH13}} contains mostly horizontal text, with some text slightly oriented.
The dataset has been widely adopted for evaluating text detection methods previously.
It consists of 229 training images and 233 testing images.

\subsection{Implementation Details}
Our network is pre-trained on SynthText and finetuned on real datasets (specified later).
It is optimized by the standard SGD algorithm with a momentum of $0.9$.
For both pretraining and finetuning, images are resized to $384\times384$ after random cropping.
Since our model is fully-convolutional, we can train it on a certain size and apply it to other sizes during testing.
Batch size is set to $32$.
In pretraining, the learning is set to $10^{-3}$ for the first $60$k iterations, then decayed to $10^{-4}$ for the rest $30$k iterations.
During finetuning, the learning rate is fixed to $10^{-4}$ for $5$-$10$k iterations.
The number of finetuning iterations depends on the size of the dataset.

Due to the precision-recall tradeoff and the difference between evaluation protocols across datasets, we choose the best thresholds $\alpha$ and $\beta$ to optimize f-measure.
Except for IC15, the thresholds are chosen separately on different datasets via a grid search with $0.1$ step on a hold-out validation set.
IC15 does not offer an offline evaluation script, so the only way for us is to submit multiple results to the evaluation server.

Our method is implemented using TensorFlow~\cite{tensorflow2015-whitepaper} r0.11.
All the experiments are carried out on a workstation with an Intel Xeon 8-core CPU (2.8 GHz), 4 Titan X Graphics Cards, and 64GB RAM.
Running on 4 GPUs in parallel, training a batch takes about 0.5s.
The whole training process takes less than a day.

\subsection{Detecting Oriented English Text}
First, we evaluate SegLink on IC15.
The pretrained model is finetuned for $10$k iterations on the training dataset of IC15.
Testing images are resized to $1280 \times 768$.
We set the thresholds on segments and links to $0.9$ and $0.7$, respectively.
Performance is evaluated by the official central submission server (\url{http://rrc.cvc.uab.es/?ch=4}).
In order to meet the requirements on submission format, the output oriented rectangles are converted into quadrilaterals.

\begin{table} 
  \caption{Results on ICDAR 2015 Incidental Text}
  \vspace{0.5em}
  \centering
  \small
  \label{tbl:ic15-results}
    \begin{tabular}{|l|c|c|c|}
    \hline
    \textbf{Method} & \textbf{Precision} & \textbf{Recall} & \textbf{F-measure}\tabularnewline
    \hline
    \hline
    HUST\_MCLAB & 47.5 & 34.8 & 40.2\tabularnewline
    \hline
    NJU\_Text & 72.7 & 35.8 & 48.0\tabularnewline
    \hline
    StradVision-2 & \textbf{77.5} & 36.7 & 49.8\tabularnewline
    \hline
    MCLAB\_FCN~\cite{cvpr/ZhangZSYLB16} & 70.8 & 43.0 & 53.6\tabularnewline
    \hline
    CTPN~\cite{eccv/TianHHH016} & 51.6 & 74.2 & 60.9\tabularnewline
    \hline
    Megvii-Image++ & 72.4 & 57.0 & 63.8\tabularnewline
    \hline
    Yao \emph{et al.}~\cite{corr/YaoBSZZC16} & 72.3 & 58.7 & 64.8\tabularnewline
    \hline
    \textbf{SegLink} & 73.1 & \textbf{76.8} & \textbf{75.0}\tabularnewline
    \hline
    \end{tabular}
\end{table}

\begin{figure*}[t]
  \centering
  \includegraphics[width=0.85\linewidth]{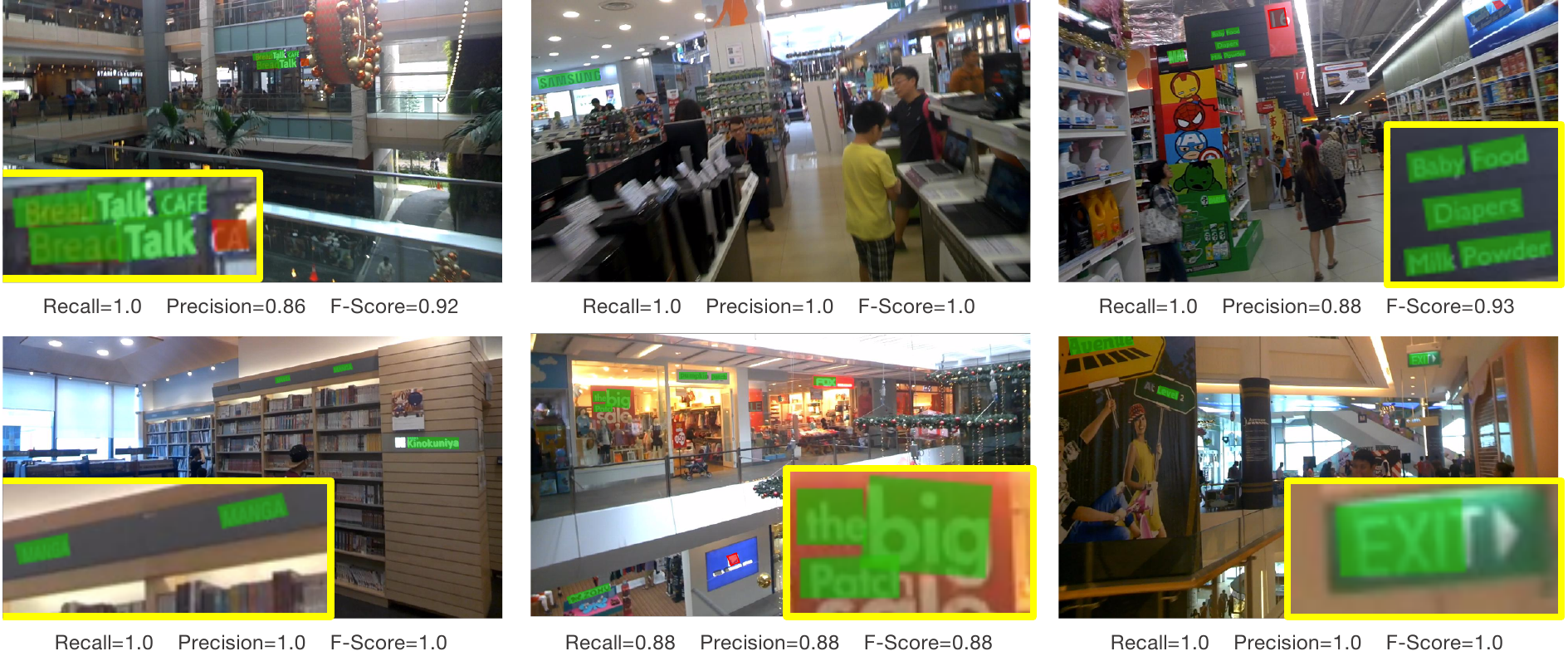}
  \caption{\textbf{Example Results on IC15.} Green regions are correctly detected text regions. Red ones are either false positive or false negative. Gray ones are detected but neglected by the evaluation algorithm. Visualizations are generated by the central submission system. Yellow frames contain zoom-in image regions. }
  \label{fig:ic15-examples}
\end{figure*}

Table~\ref{tbl:ic15-results} lists and compares the results of the proposed method and other state-of-the-art methods.
Some results are obtained from the online leaderboard.
SegLink outperforms the others by a large margin.
In terms of f-measure, it outperforms the second best by 10.2\%.
Considering that some methods have close or even higher precision than SegLink, the improvement mainly comes from the recall.
As shown in Fig.~\ref{fig:ic15-examples}, our method is able to distinguish text from very cluttered backgrounds.
In addition, owing to its explicit link prediction, SegLink correctly separates words that are very close to each other.

\subsection{Detecting Multi-Lingual Text in Long Lines} \label{sec:exp-td500}
We further demonstrate the ability of SegLink to detect long text in non-Latin scripts.
TD500 is taken as the dataset for this experiment, as it consists of oriented and multi-lingual text.
The training set of TD500 only has 300 images, which are not enough for finetuning our model.
We mix the training set of TD500 with the training set of IC15, in the way that every batch has half of its images coming from each dataset.
The pretrained model is finetuned for $8$k iterations.
The testing images are resized to $768\times 768$.
The thresholds $\alpha$ and $\beta$ are set to $0.9$ and $0.5$ respectively.
Performance scores are calculated by the official development toolkit.

\begin{table}[hbt] 
  \caption{Results on MSRA-TD500}
  \vspace{0.5em}
  \centering
  \small
  \label{tbl:td500-results}
  \begin{tabular}{|l|c|c|c|c|}
  \hline
  \textbf{Method} & \textbf{Precision} & \textbf{Recall} & \textbf{F-measure} & \textbf{FPS}\tabularnewline
  \hline
  \hline
  Kang \emph{et al.}~\cite{cvpr/KangLD14} & 71 & 62 & 66 & -\tabularnewline
  \hline
  Yao \emph{et al.}~\cite{cvpr/YaoBLMT12} & 63 & 63 & 60 & 0.14\tabularnewline
  \hline
  Yin \emph{et al.}~\cite{pami/YinPZH15} & 81 & 63 & 74 & 0.71\tabularnewline
  \hline
  Yin \emph{et al.}~\cite{pami/YinYHH14} & 71 & 61 & 65 & 1.25\tabularnewline
  \hline
  Zhang \emph{et al.}~\cite{cvpr/ZhangZSYLB16} & 83 & 67 & 74 & 0.48\tabularnewline
  \hline
  Yao \emph{et al.}~\cite{corr/YaoBSZZC16} & 77 & \textbf{75} & 76 & $\sim$1.61\tabularnewline
  \hline
  \textbf{SegLink} & \textbf{86} & 70 & \textbf{77} & \textbf{8.9}\tabularnewline
  \hline
  \end{tabular}
\end{table}

\begin{figure*}[t]
  \centering
  \includegraphics[width=0.85\linewidth]{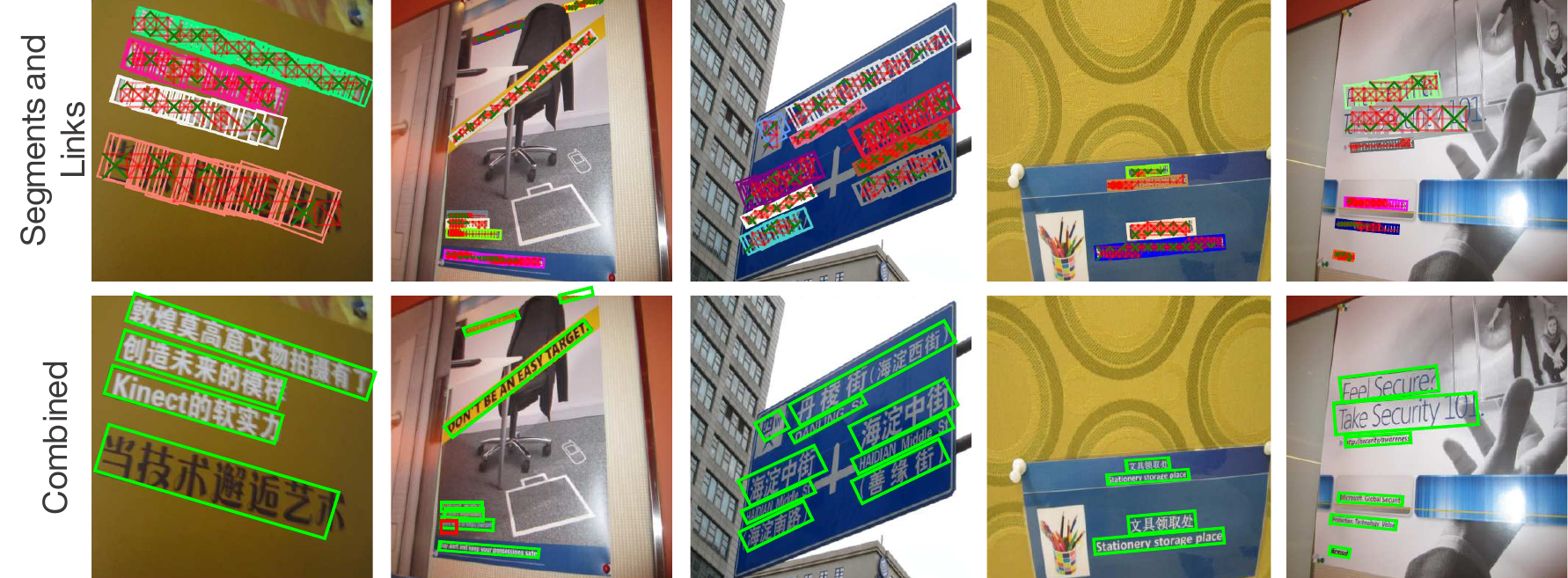}
  \caption{\textbf{Example Results on TD500.} The first row shows the detected segments and links. The within-layer and cross-layer links are visualized as red and green lines, respectively. Segments are shown as rectangles in different colors, denoting different connected components. The second row shows the combined boxes.}
  \label{fig:td500-examples}
\end{figure*}

According to Table~\ref{tbl:td500-results}, SegLink achieves the highest scores in terms of precision and f-measure.
Benefiting from its fully-convolutional design, SegLink runs at 8.9 FPS, a much faster speed than the others.
SegLink also enjoys simplicity.
The inference process of SegLink is a single forward pass in the detection network,
while the previous methods \cite{cvpr/YaoBLMT12, pami/YinYHH14, cvpr/ZhangZSYLB16} involve sophisticated rule-based grouping or filtering steps.

TD500 contains many long lines of text in mixed languages (English and Chinese).
Fig.~\ref{fig:td500-examples} shows how SegLink handles such text.
As can be seen, segments and links are densely detected along text lines.
They result in long bounding boxes that are hard to obtain from a conventional object detector.
Despite the large difference in appearance between English and Chinese text, SegLink is able to handle them simultaneously without any modifications in its structure.

\subsection{Detecting Horizontal Text}
Lastly, we evaluate the performance of SegLink on horizontal-text datasets.
The pretrained model is finetuned for $5$k iterations on the combined training sets of IC13 and IC15.
Since the most text in IC13 has relatively larger sizes, the testing images are resized to $512 \times 512$.
The thresholds $\alpha$ and $\beta$ are set to $0.6$ and $0.3$, respectively.
To match the submission format, we convert the detected oriented boxes into their axis-aligned bounding boxes.

Table~\ref{tbl:ic13-results} compares SegLink with other state-of-the-art methods.
The scores are calculated by the central submission system using the ``Deteval'' evaluation protocol.
SegLink achieves very competitive results in terms of f-measure.
Only one approach \cite{eccv/TianHHH016} outperforms SegLink in terms of f-measure.
However, \cite{eccv/TianHHH016} is mainly designed for detecting horizontal text and is not well-suited for oriented text.
In terms of speed, SegLink runs at over 20 FPS on $512\times 512$ images, much faster than the other methods.

\begin{table}[hbt] 
  \caption{\textbf{Results on IC13}. P, R, F stand for precision, recall and f-measure respectively. *These methods are only evaluated under the ``ICDAR 2013'' evaluation protocol, the rest under ``Deteval''. The two protocols usually yield very close scores.}
  \vspace{0.5em}
  \centering
  \small
  \label{tbl:ic13-results}
  \begin{tabular}{|l|c|c|c|c|}
  \hline
  \textbf{Method} & \textbf{P} & \textbf{R} & \textbf{F} & \textbf{FPS}\tabularnewline
  \hline
  \hline
  Neumann \emph{et al.}~\cite{icdar/NeumannM15}$^{*}$ & 81.8 & 72.4 & 77.1 & 3\tabularnewline
  \hline
  Neumann \emph{et al.}~\cite{pami/NeumannM16}$^{*}$ & 82.1 & 71.3 & 76.3 & 3\tabularnewline
  \hline
  Busta \emph{et al.}~\cite{iccv/BustaNM15}$^{*}$ & 84.0 & 69.3 & 76.8 & 6\tabularnewline
  \hline
  Zhang \emph{et al.}~\cite{cvpr/ZhangSYB15} & 88 & 74 & 80 & \textless{}0.1\tabularnewline
  \hline
  Zhang \emph{et al.}~\cite{cvpr/ZhangZSYLB16} & 88 & 78 & 83 & \textless{}1\tabularnewline
  \hline
  Jaderberg \emph{et al.}~\cite{ijcv/JaderbergSVZ16} & 88.5 & 67.8 & 76.8 & \textless{}1\tabularnewline
  \hline
  Gupta \emph{et al.}~\cite{cvpr/GuptaVZ16} & 92.0 & 75.5 & 83.0 & 15\tabularnewline
  \hline
  Tian \emph{et al.}~\cite{eccv/TianHHH016} & \textbf{93.0} & \textbf{83.0} & \textbf{87.7} & 7.1\tabularnewline
  \hline
  \textbf{SegLink} & 87.7 & \textbf{83.0} & 85.3 & \textbf{20.6}\tabularnewline
  \hline
  \end{tabular}
\end{table}

\begin{figure}[h]
  \centering
  \includegraphics[width=\linewidth]{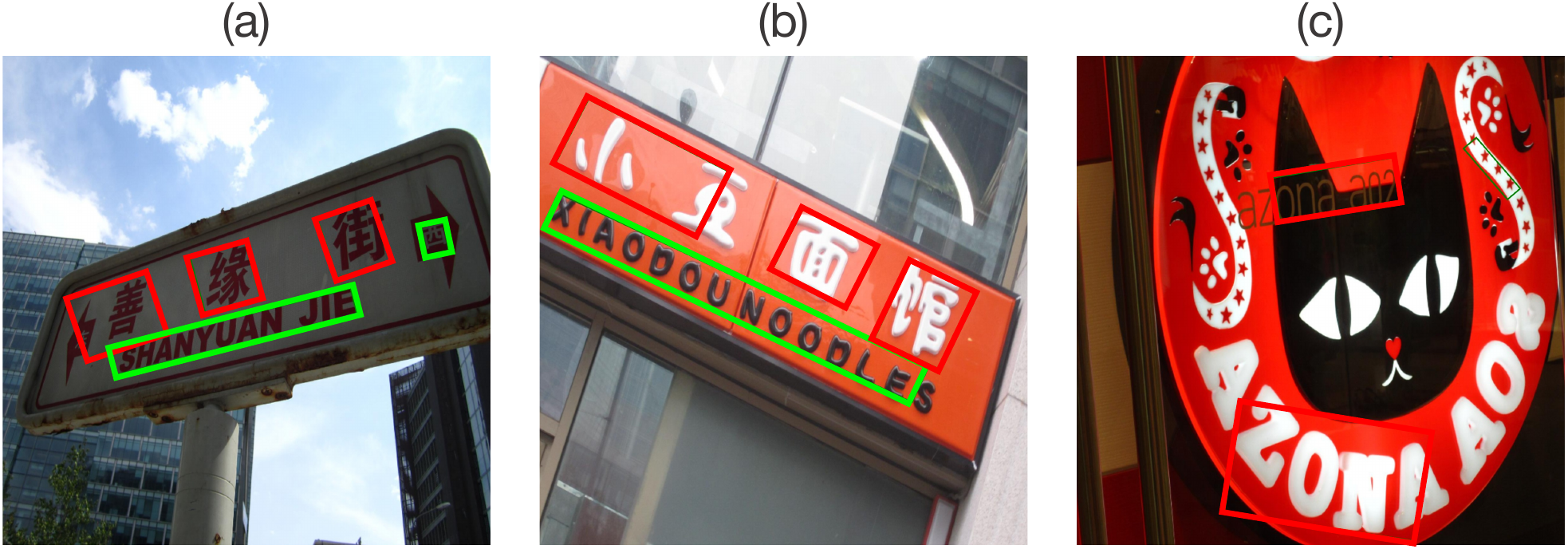}
  \caption{\textbf{Failure cases on TD500}. Red boxes are false positives. (a)(b) SegLink fails to link the characters with large character spacing. (c) SegLink fails to detect curved text.}
  \label{fig:fail-cases}
\end{figure}

\subsection{Limitations}
A major limitation of SegLink is that two thresholds, $\alpha$ and $\beta$, need to be set manually.
In practice, the optimal values of the thresholds are found by a grid search.
Simplifying the parameters would be part of our future work.
Another weakness is that SegLink fails to detect text that has very large character spacing.
Fig.~\ref{fig:fail-cases}.a,b show two such cases.
The detected links connect adjacent segments but fail to link distant segments.

Fig.~\ref{fig:fail-cases}.c shows that SegLink fails to detect text of curved shape.
However, we believe that this is not a limitation of the segment linking strategy, but the segment combination algorithm, which can only produce rectangles currently.

\section{Conclusion}
We have presented SegLink, a novel text detection strategy implemented by a simple and highly-efficient CNN model.
The superior performance on horizontal, oriented, and multi-lingual text datasets well demonstrate that SegLink is accurate, fast, and flexible.
In the future, we will further explore its potentials on detecting deformed text such as curved text.
Also, we are interested in extending SegLink into a end-to-end recognition system.

\section*{Acknowledgment}

This work was supported in part by National Natural Science Foundation of China (61222308 and 61573160), a Google Focused Research Award, AWS Cloud Credits for Research, a Microsoft Research Award and a Facebook equipment donation. The authors also thank China Scholarship Council (CSC) for supporting this work.

{\small
\bibliographystyle{ieee}
\bibliography{references}
}

\end{document}